\pdfoutput=1

\documentclass[11pt]{article}

\usepackage[]{emnlp2021}

\usepackage{times}
\usepackage{latexsym}

\usepackage[T1]{fontenc}

\usepackage[utf8]{inputenc}

\usepackage{microtype}

\usepackage{subcaption}
\usepackage{graphicx}

\title{Pay Attention When Required}

\author{
Swetha Mandava \\
  \texttt{smandava@nvidia.com} \\ \And
Szymon Migacz \\
  \texttt{smigacz@nvidia.com} \\ \And
Alex Fit-Florea \\
\texttt{afitflorea@nvidia.com}\\}

\date{}

\begin{document}
\maketitle
\begin{abstract}
Transformer-based models consist of interleaved feed-forward blocks - that capture content meaning, and relatively more expensive self-attention blocks - that capture context meaning. In this paper, we explored trade-offs and ordering of the blocks to improve upon the current Transformer architecture and proposed PAR Transformer. It needs 35\% lower compute time than Transformer-XL achieved by replacing ~63\% of the self-attention blocks with feed-forward blocks and retains the perplexity on WikiText-103 language modeling benchmark. We further validated our results on text8 and enwiki8 datasets, as well as on the BERT model.

\end{abstract}

\section{Introduction}

The seminal work in ~\citep{attention} introduced the Transformer architecture. Since its introduction, it profoundly influenced algorithms for Question Answering, Text Classification, Translation, Language Modeling, and practically all of the Natural Language Processing tasks. A transformer layer consists of interleaved self attention and feed forward blocks and is used in state of the art models, like Transformer-XL ~\citep{TXL}, BERT ~\citep{bert}, Megatron ~\citep{megatron}, and other large-scale language models.

As corresponding model sizes and compute requirements continue to become increasingly more demanding, it becomes important to optimize the Transformer-based architectures, for both financial as well as environmental reasons ~\citep{strubell2019energy}. Several optimization approaches, that used pruning ~\citep{droppingheads}, and distillation ~\citep{distilbert,jiao2020tinybert,wang2020minilm}, were able to achieve better run-time performance for an accuracy trade-off. 

Our optimization approach investigates the trade-off between the self-attention and feed-forward building blocks. We start with the intuition that attention blocks provides context meaning while being comparatively more expensive, and feed-forward blocks provide content meaning. We then ask the fundamental questions of what are the saturation points when using one block versus the other, and how accuracy depends on the relative number of blocks of each type as well as on their ordering. To answer these questions, we employed architecture search. 

While recent works such as ~\citep{fbnet,fbnet2,darts} explored using differential neural architecture search for designing ConvNets automatically to significantly improve accuracies and/or latencies, similar work for transformer models is limited. Recent works, however, explored using random search ~\citep{SandwichTransformer} and evolutionary search ~\citep{hwawaretrans,evolvedtrans} for designing transformer models. However, even with a search space of three options per layer (Self Attention, Feed Forward, Identity), the design space becomes intractable for 32 layers as it is combinatorial (=$3^{32}$). For this reason, we explored the use of differential neural architecture search that has linear complexity to redesign the transformer architecture in this paper.

In order to analyze the transformer architecture, we studied the search on Transformer-XL Base with the WikiText-103 dataset. The analysis of the resulting optimal architectures, highlights two fundamental rules:

\begin{enumerate}
    \item Self-attention layers are necessary only among the former two-thirds layers of the network
    \item The total number of layers to self-attention layers ratio of p:1 is sufficient, with p=5 being optimal for Transformer-XL.
\end{enumerate}

We propose \textbf{P}ay \textbf{A}ttention when \textbf{R}equired Transformer (or \textbf{PAR} Transformer), a new family of models based on the above two design rules, that uses ~63\% fewer self-attention blocks while retaining test accuracies. Further, we validated that our hypothesis generalizes to different datasets (text8, enwiki8) as well as transformer models (PAR BERT) for different tasks (Question Answering, Sentiment Analysis, Semantic Textual Similarity).

\section{Optimal Design Rules for Transformers}

Our baseline, the Transformer-XL model, has equal number of self-attention and feed forward blocks in an interleaved design pattern as visualized in Figure \ref{fig:model_lat}. Sandwich transformers ~\citep{SandwichTransformer}, also keeps an equal number of self-attention and feed forward blocks but are designed using a sandwich coefficient $k$ instead of having a simple interleaved design pattern. They have the first $k$ sublayers consisting of self-attention, the last $k$ sublayers consisting of feed forward layers with both sandwiched between the classic interleaving pattern of self-attention and feed forward blocks. This design pattern was found by conducting a series of random search experiments with constraints to keep the number of parameters constant. 

In this section, we attempt to optimize the transformer architecture by relaxing the above search constraints. We employ differential neural architecture search and allow it to select one of the three options - Self Attention, Feed Forward, or Identity -- for each of the layers.

\begin{figure}[h]
    \centering
    \includegraphics[width=0.48\textwidth]{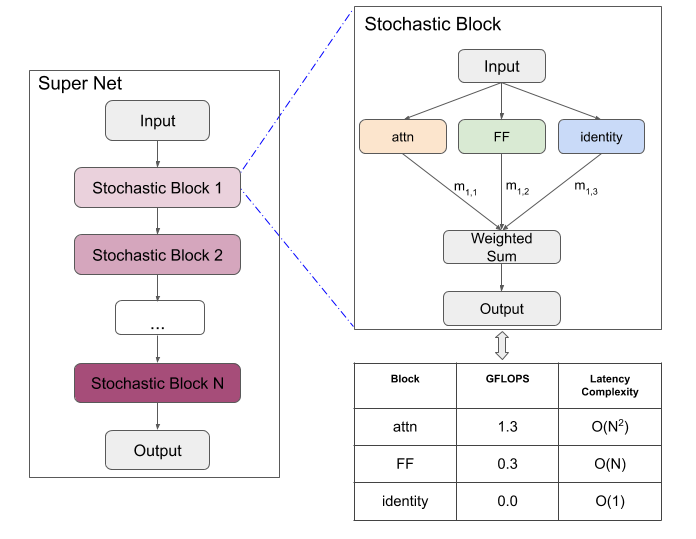}
    \captionsetup{justification=centering}
    \caption{Composition of Super Net as a linear combination of search blocks along with the block cost.\\
    GFLOPS computed for inference with $tgt\_len$ = 64, $mem\_len$ = 640, $batch\_size$ = 1.  \\
    Latency complexity with respect to sequence length.}
    \label{fig:supernet}
\end{figure}  

\subsection{Search Space}

Is interleaved attention and feed forward layers in a transformer really the optimal design pattern? Can we get the same results for smaller, faster or imbalanced networks? In order to test these questions, we used a very simple search space that would allow us to do so, consisting of identity block, feed forward block and self-attention block that modify the input sequence $X$ as follows:

\begin{equation}
LN(X)= LayerNorm(X)
\end{equation}%

\begin{equation}
F_{attn}(X)=Self\mbox{-}Attention(LN(X)) + X
\end{equation}%

\begin{equation}
F_{FF}(X)=Feed\mbox{-}Forward(LN(X) ) + X
\end{equation}%

\begin{equation}
F_{identity}(X)=X
\end{equation}%

The output of each layer $l$ can be computed using equation  \ref{eqn:layer_output} where $i$ is the block choice and $m_{l,i}$ is a probability distribution computed by a Gumbel Softmax function ~\citep{gumbel_softmax1,gumbel_softmax2} on all the choices in a layer from the search space. Once trained, $m_{l,i}$ allows us to study the optimal models. For example, if identity block is the most possible block on a layer, we can hypothesize that there is no benefit from a deeper network. Similarly, the model can also pick different design patterns and faster models.

\begin{equation}
\sum_{i \in (attn, FF, identity)} m_{l,i}=1
\label{eqn:arch_params}
\end{equation}%

\begin{equation}
X^{l}=\sum_{i \in (attn, FF, identity)} m_{l,i}. F^{l}_{i} (X^{l-1})
\label{eqn:layer_output}
\end{equation}%

Since the output at each layer is a linear combination of individual search blocks in that layer, the search cost is linear with respect to the number of blocks in the search space. Since the search also consists of training only one supernet consisting of all the search blocks, it is orders of magnitude faster than RL based search algorithms ~\citep{rl1,rl2} that rely on training individual combinations of search blocks. For our choice of supernet, the search cost was $<2 \times$ the training cost of the baseline. All of our experiments use the same model architecture parameters as the baseline from Table \ref{table:joc_txl} unless otherwise mentioned.

\subsection{Search Algorithm and Experiments} \label{observations}

In order to explore design paradigms of a transformer architecture, we use differential neural architecture search, similar to FBNet Algorithm ~\citep{fbnet}, formulated as a two stage search shown in equation \ref{eq:search_fn} where the goal is to find the architecture $a$ within a search space $A$, and weights $w_a$ that minimizes the loss function $L_{a, w_a}$ or the cross entropy loss. Within the architecture phase, architecture parameters $m_{l,i}$ are tuned and within the weight phase, weight parameters of the individual search blocks are tuned, to minimize the loss.

\begin{equation}\label{eq:search_fn}
a,w_a=min_{a \in A} min_{w_a} E_{a \sim P_{\theta}} \big\{ L_{a, w_a} \big\}
\end{equation}%



\begin{figure}
\centering
\includegraphics[width=0.48\textwidth]{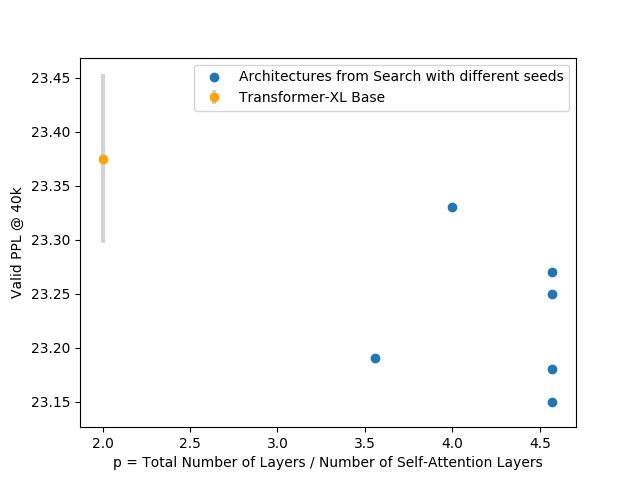}
\caption{Perplexities on WikiText-103 dev set as a function of number of self-attention blocks, for Total Number of Layers = 32. Architectures from search are obtained from 6 random seeds, and re-trained from scratch for 40k iterations. Transformer-XL base indicates mean +- std perplexity from 6 random seeds}
\label{fig:attn_num}
\end{figure}

\begin{figure}
\centering
\includegraphics[width=0.48\textwidth]{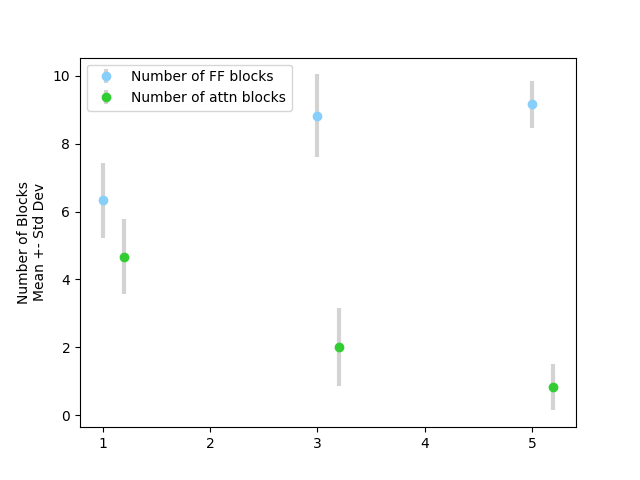}
\caption{Analysis of number of blocks within each slice of the model for architectures from search, with 6 random seeds}
\label{fig:attn_loc}
\end{figure}

We run the neural architecture search described above on $16 \times 2=32$ layers for the WikiText-103 ~\citep{wikitext103} dataset. On each layer, the search algorithm is able to choose between a feed-forward, self-attention or identity blocks. We run the search algorithm with batch size=128, architecture update lr=1e-2, weight decay for architecture parameters=5e-4, weight update lr=1e-2, weight decay for block weights=1e-4. We initialize the architecture parameters uniformly and keep architecture parameters constant for the first 10k iterations and then perform architecture update for 20\% of an epoch from there on for 40k iterations. We train till the architecture converges i.e. does not change in 75\% of the architecture tuning stage.

The differential architecture search produces probability distributions $m_{l,i}$ representing the likelihood of block $i$ being the optimal choice for layer $l$. At each layer, the most probable block is selected. We repeated this search process for 6 random seeds, and retrained the searched models from scratch. Analyzing these search architectures and their performance revealed interesting properties.

We first observe that total number of layers to self-attention layers ratio is much higher than 2:1. In Figure \ref{fig:attn_num}, we see that we are able to achieve lower perplexities than the baseline by architectures that use fewer self-attention blocks. We then analyzed where these self-attention blocks are located within the model network. To do this, we split the model into three slices and counted the number of self-attention or feed forward blocks in each of these slices. In Figure \ref{fig:attn_loc}, we see that the majority of self-attention blocks are in the former two-third layers of the network ie the mean number of self-attention blocks in the final third of the layers is $<1$.

Previous research ~\citep{darksecretsofbert,collaborate} on transformer layers also indicate that attention layers are severely over parametrized and that they are more prevalent in the beginning of a network ~\citep{SandwichTransformer}. Our aim is to quantify how much and when they are needed and to formalize that in a set of rules.

\subsection{Formalizing Design Rules For Transformers} \label{design_rules}

While this process of searching for an architecture and re-training the optimal architecture from scratch for a particular dataset and a particular model can be employed, it is expensive. The scope of this paper is to understand generalizable design rules that can be applied to different transformer models and datasets. To this end, we attempt to hypothesize optimal design rules based on the observations in section \ref{observations} and validate them in the following sections. 

\begin{figure}
\centering
\includegraphics[width=0.48\textwidth]{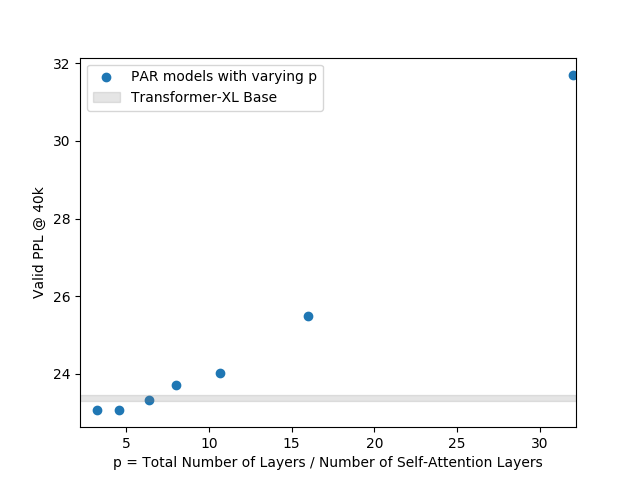}
\caption{Perplexities on WikiText-103 dev set with respect to PAR coefficient p, for total number of layers = 32. Horizontal line indicates mean +- std perplexity of our baseline, Transformer-XL Base, from 6 random seeds}
\label{fig:num_attn_plot}
\end{figure}


Our observations in previous sections motivate us to design a family of PAR Transformer models that have fewer self-attention blocks that are positioned in the former two-third layers of the network. A PAR transformer is now, formalized as being based on the following two design rules: 

\begin{enumerate}
    \item Self-attention layers are only needed among the former two-third layers of the network
    \item Total number of layers to self-attention layers ratio of p:1 is sufficient.
\end{enumerate}

We can now attempt to design optimized transformer models manually based on these design rules. For example, to design a transformer architecture with 32 layers, for a PAR Coefficient of p=5, we use $6 \approx (32/5)$ self-attention layers. These self-attention layers are placed uniformly within the first $21 \approx (2 * 32 / 3)$ layers. 

We train PAR transformers for various $p > 2$ (to use fewer self-attention blocks than our baseline). Figure \ref{fig:num_attn_plot} shows the performance as a function of PAR coefficient. Of those models, PAR coefficient of 5 is sufficient to match the accuracy of our baseline. We identify this 32 layer, p=5 PAR model as PAR Transformer Base.

\begin{figure}
\centering
\includegraphics[width=0.48\textwidth]{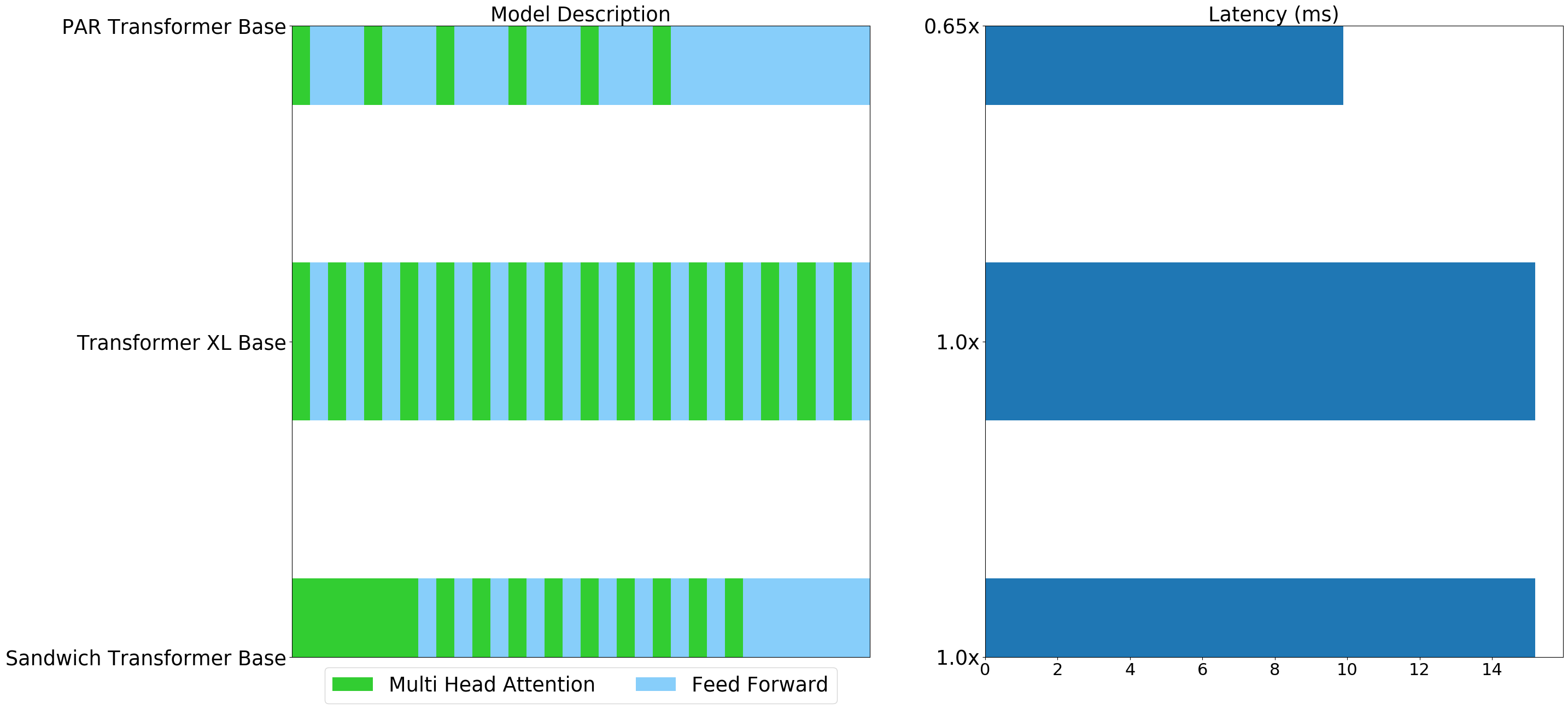}
\caption{Comparison of Model Architecture and Latency on A100}
\label{fig:model_lat}
\end{figure}

We also observe that while self-attention blocks are essential for contextual meaning, the need is saturated fairly quickly. The advantage of replacing self-attention blocks with feed-forward blocks is the significant latency benefit we obtain, as shown in Figure \ref{fig:model_lat}. The benefits are even more pronounced with higher sequence lengths as self-attention and feed forward blocks have $O(N^{2})$ and $O(N)$ complexities per-layer with respect to sequence lengths, respectively.


\begin{table}[]
    \centering
    {\resizebox{0.48\textwidth}{!}{
     \begin{tabular}{cccc} 
     \hline
    \textbf{Model} & \textbf{Architecture} & \textbf{Latency} & \textbf{PPL} \\ 
    & & \textbf{on A100 (ms)} & \\ [0.5ex] 
     \hline\hline
     Transformer-XL Base & (sf)x16 & 15.2 & 22.7 \\ [1ex] 
     Sandwich Transformer Base & (s)x6 (sf)x10 (f)x6 & 15.2 & 22.6 \\ [1ex]
     PAR Transformer Base & (sfff)x6 (f)x8 & 9.9 & 22.7 \\ [1ex] 
     \hline\hline
    \end{tabular}}}
    \captionsetup{justification=centering}
    \caption{Latency and Perplexity (PPL) of Transformer-XL Base models on WikiText-103 dataset.}
    \label{table:wiki103_results}
\end{table}

\begin{table*}[hbt!]
    \centering
    {\resizebox{0.98\textwidth}{!}{
     \begin{tabular}{ccccc} 
     \hline
    \textbf{Dataset} & \textbf{Model} & \textbf{Architecture} & \textbf{Latency on A100 (ms)} & \textbf{bpc / PPL} \\ [0.5ex] 
     \hline\hline
    WikiText-103 & Transformer-XL Large & (sf)x18 & 18.9 & 18.4 \\ [1ex] 
     & Sandwich Transformer Large & (s)x6 (sf)x12 (f)x6 & 18.9 & 18.2 \\ [1ex]
     & PAR Transformer Large & (sfff)x7 (f)x8 & 13.4 & 18.4 \\ [1ex] 
     \hline\hline
     enwiki8 & Transformer-XL 24B & (sf)x12 & 12.5 & 1.10 \\ [1ex] 
     & Sandwich Transformer 24B & (s)x5 (sf)x7 (f)x5 & 12.5 & 1.10 \\ [1ex]
     & PAR Transformer 24B & (sff)x5 (f)x9 & 8.4 & 1.11 \\ [1ex] 
     \hline\hline
    text8 & Transformer-XL 24B & (sf)x12 & 12.5 & 1.18 \\ [1ex] 
     & Sandwich Transformer 24B & (s)x3 (sf)x9 (f)x3 & 12.5 & 1.18 \\ [1ex]
     & PAR Transformer 24B & (sff)x5 (f)x9 & 8.4 & 1.18 \\ [1ex] 
     \hline
    \end{tabular}}}
    \captionsetup{justification=centering}
    \caption{Bits Per Character (bpc) on enwiki8 and text8 and Perplexity (PPL) on WikiText-103 for Transformer-XL models.}
    \label{table:otherdata_results}
\end{table*}

\begin{table*}[]
    \centering
    {\resizebox{0.98\textwidth}{!}{
     \begin{tabular}{cccccc} 
     \hline
    \textbf{Model} & \textbf{Architecture} & \textbf{Latency on A100 (ms)} & \textbf{SQuAD v1.1} & \textbf{SST-2} & \textbf{MRPC} \\ [0.5ex] 
     \hline\hline
     DistilBERT* & (sf)x6  & $5.3^{+}$ & 86.9 & 91.3 & 87.5 \\ [1ex]
     BERT Base & (sf)x12 & 8.6 & 88.4 & 91.5 & 88.7 \\ [1ex] 
     PAR BERT Base & (sff)x5 (f)x9 & 5.7 & 87.4 & 91.6 & 89.2 \\ [1ex] 
     \hline
    \end{tabular}}}
    \captionsetup{justification=centering}
    \caption{Experimental results of PAR BERT in comparison to BERT Base and DistilBERT. 
    \hspace{\textwidth}F1 score for SQuAD v1.1 and accuracy for SST-2 and MRPC reported from a median of 5 runs on dev sets.
    \hspace{\textwidth}Reported Latency for SQuAD inference.
    \hspace{\textwidth} * indicates originally published results.
    \hspace{\textwidth} + indicates estimated latency as 61\% of Bert Base based on DistilBERT paper}
    \label{table:bert_results}
\end{table*}

\section{Experiments}

In this section, we review our results on PAR Transformer Base with WikiText-103 with respect to state of the art transformer architectures. We further validate that the PAR design rules generalize to other Transformer-XL models (Large, 24B) and to other datasets (enwiki8, text8). We also validate our design rules on BERT models with PAR BERT. 

All the models are based on the same code base for training, for an apples-to-apples comparison. We used NVIDIA A100 40GB Tensor Core GPUs for our experiments. The Architecture column explains the composition of the model, with \textbf{s} indicating a self-attention block and \textbf{f} indicating a feed-forward block. PAR model architectures are modelled using PAR design rules underlined in section \ref{design_rules} for PAR coefficient p=5. Sandwich Transformer model architectures are based on optimal sandwich coefficients specified for each dataset when indicated in their paper. Latencies indicate inference latencies for batch size 1 as is standard. We see similar performance benefits while training as well.

\subsection{PAR Transformer}

We compare the performance of our PAR Transformer on WikiText 103 dataset in Table \ref{table:wiki103_results}. WikiText 103 language modelling dataset consists of over 100 million tokens from articles on Wikipedia. It is well suited for testing long term dependencies as it is composed of full articles and with original case, punctuation and numbers. The only difference in the model architectures is the ordering and composition of layers, as visualized in Figure \ref{fig:model_lat} and listed under the Architecture column. 

The Transformer-XL Base code is based on the code published by the authors from the Transformer-XL paper but modifies hyper parameters as described in Table \ref{table:joc_txl} for better hardware utilization in base model. Inference latencies are computed using $tgt\_len$ = 64, $mem\_len$ = 640 and $clamp\_len$ = 400. We validate that we are able to obtain the same perplexities with 0.65x the cost in terms of latency. 

We further validate that our hypothesis generalizes to the PAR Transformer Large Model in Table \ref{table:otherdata_results}, by maintaining the perplexities with 0.7x the cost. The Large model uses 36 layers, $d\_model$ = 1024, $d\_head$ = 64, $n\_head$ = 16, $d\_inner$ = 4096, $tgt\_len = mem\_len = $384, $batch size$ = 128 for training and $tgt\_len$ = 128, $mem\_len$ = 1600, $clamp\_len$ = 1000 for evaluation.

In order to showcase generalizability over different datasets, we validate our results on enwiki8 and text8 datasets ~\citep{textenwiki} in Table \ref{table:otherdata_results}. Enwiki8 consists of 100M bytes of unprocessed Wikipedia text whereas text8 contains 100M characters of preprocessed Wikipedia text. We reuse the same model hyperparameters as in Table \ref{table:joc_txl} with 24 layers. In addition, we use $tgt\_len$ = $mem\_len$ = 512 for training and $tgt\_len$ = 128, $mem\_len$ = 2052, $clamp\_len$ = 820 for evaluation.

\begin{figure}
\centering
\includegraphics[width=0.48\textwidth]{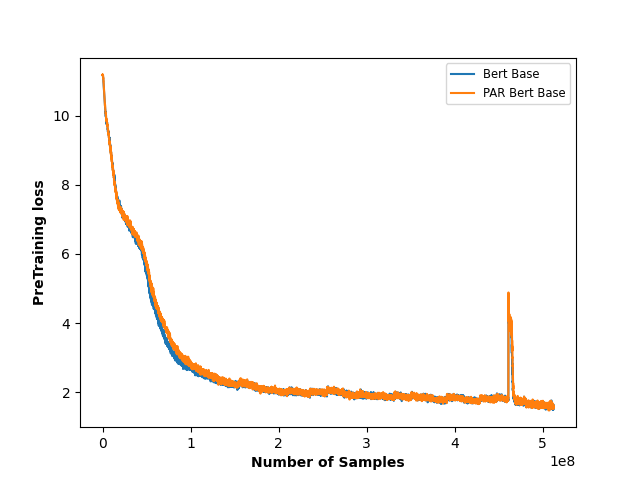}
\caption{Pretraining Loss curve using NVLamb Optimizer for BERT Base and PAR BERT Base models}
\label{fig:bert_pretraining}
\end{figure}

\subsection{PAR BERT}

We further study the effect of PAR design rules on BERT models by pre-training on Wikipedia+Books datasets ~\citep{bookscorpus} using the NVLAMB optimizer ~\citep{nvlamb} in two phases. Phase 1 is trained with a sequence length of 128, with a batch size of 64k for 7038 steps and phase 2 is trained with a sequence length of 512, with a batch size of 32k for 1563 steps. 

Our pre-training loss curves in Figure \ref{fig:bert_pretraining} highlight the on-par performance of PAR BERT and BERT Base with a fraction of the self-attention blocks. We see in Table \ref{table:bert_results} that using the same architectural design rules results in a 1\% accuracy drop on SQuAD v1.1 fine-tuning task even though the pre-training loss and accuracy on MRPC and SST-2 are on track. We hypothesize that tuning PAR coefficient specifically might help bridge the gap. However, incorporating the two stage optimization process (pre-training followed by fine-tuning) that is inherent to BERT and other such language models into architecture tuning is a future research problem.

It is, however, interesting to note that PAR BERT has comparable latencies with respect to DistilBERT even though it uses twice as many layers. PAR BERT also outperforms DistilBERT while having a much simpler training paradigm. Nevertheless, we note that pruning, quantization and distillation are orthogonal to the present work and could be used in conjunction.

\section{Conclusion}

We used differential neural architecture search to study patterns in the ordering of transformer model sub-layers and made two key observations. One, that we only need attention layers in the former part of the network and two, that we need 63\% fewer attention layers to retain the model accuracy. Even though we studied the search results specifically on Transformer-XL Base for the WikiText-103 dataset, the same observations were valid for other transformer models and datasets as well. 

We proposed PAR Transformer that achieves 35\% lower latency and validated accuracy on enwiki8 and text8 datasets as well as on 24B + Large variations. We also validated our results on SQuAD v1.1, MRPC and SST-2 with PAR BERT Base model. 

In this paper, we used differential neural architecture search to make optimal composition of transformer architectures explainable. It provides an avenue for automatic design of optimized model families.

\nocite{*}
\bibliographystyle{acl_natbib}
\bibliography{custom}

\appendix
\section{Appendix}
\setcounter{table}{0}
\renewcommand{\thetable}{A\arabic{table}}
\subsection{Hyper parameter changes to Model}

Our Transformer-XL (Base, 24B) baselines are based on the code base published by the authors of the Transformer-XL paper but uses a modified set of model hyper parameters. Our modifications were made to achieve better hardware utilization and to take advantage of Tensor Cores, most commonly by aligning certain hyper parameters with powers of two. They are described in Table \ref{table:joc_txl}. 

\begin{table}[hbt!]
    \centering
    {\resizebox{0.48\textwidth}{!}{%
    \begin{tabular}{cccc}
     \hline
    \textbf{Hyperparameter} & \textbf{Description} & \textbf{Original setting} & \textbf{Our modification} \\ [0.3ex] 
    \hline\hline
    $d_{model}$ & hidden size & 410 & 512 \\[0.3ex]
    $n_{head}$  & number of attention heads & 10  & 8   \\[0.3ex]
    $d_{head}$  & size of each attention head & 41  & 64  \\[0.3ex]
    $d_{inner}$ & hidden size in & 2100 & 2048 \\
    & fully-connected layers & & \\ [0.3ex]
    $tgt\_len$ & number of tokens & 150 & 192 \\ 
    & to predict during training & & \\[0.3ex]
    $mem\_len$ & number of tokens cached from & 150 & 192 \\
    & previous iterations during training & & \\[0.3ex]
    \hline
    \end{tabular}}}
\caption{Hyper parameter modifications made to Transformer-XL Base}
\label{table:joc_txl}
\end{table}

\subsection{\#Parameters \#Flops with respect to Latency}

Even though literature generally reports number of parameters to estimate efficiency of a model, it is too simplistic, often obscuring performance issues rather than illuminating them. We can see from Table \ref{table:flops_and_params} that \#Parameters don't actually reflect the latency. While \#FLOP count is hardware independent to its merit, latency is less abstract and more indicative of actual performance.

\begin{table}
\centering
{\resizebox{0.48\textwidth}{!}{%
\begin{tabular}{lrlrl}
     \hline
    \textbf{Model} & \textbf{Architecture} & \textbf{\#Params} & \textbf{\#GFLOPs} & \textbf{Latency} \\
    & & & & \textbf{on A100 (ms)} \\ [0.5ex] 
     \hline\hline
     Transformer-XL Base & (sf)x16 & 192M & 27 & 15.2 \\ [1ex]
     Sandwich Transformer Base & (s)x6 (sf)x10 (f)x6 & 192M & 27 & 15.2 \\[1ex]
     PAR Transformer Base & (sfff)x6 (f)x8 & 200M & 17 & 9.9 \\ [1ex] 
    \hline
    \end{tabular}}}
    \captionsetup{justification=centering}
    \caption{Flops and Parameters with respect to Latency for Base Models}
    \label{table:flops_and_params}
\end{table}

\begin{table}[hbt!]
    \centering
    {\resizebox{0.48\textwidth}{!}{%
     \begin{tabular}{ccc} 
     \hline
    \textbf{Model} & \textbf{Valid PPL} & \textbf{Valid PPL} \\
    & \textbf{@ 40k} & \textbf{@ 140k} \\
     \hline\hline
     Transformer-XL Base & 23.3 & 22.2 \\ [1ex]
     Sandwich Transformer Base & 23.4 & 22.4 \\[1ex]
     PAR Transformer Base & 23.3 & 22.4 \\ [1ex]
     \hline
    \end{tabular}}}
    \captionsetup{justification=centering}
    \caption{Latency and Valid Perplexity on WikiText-103 dataset with respect to training steps}
    \label{table:wiki103_wrt_epochs}
\end{table}

\subsection{Accuracy with respect to training steps}

Table \ref{table:wiki103_wrt_epochs} lists test perplexities at 40k and 140k iterations with a global batch size of 256. As we can see, there is little benefit in training much further after 40k iterations for the base model.

\end{document}